\documentclass[letterpaper]{article} 
\usepackage{aaai2026}  
\usepackage{times}  
\usepackage{helvet}  
\usepackage{courier}  
\usepackage[hyphens]{url}  
\usepackage{graphicx} 
\usepackage{natbib}  
\usepackage{caption} 
\usepackage{algorithm}
\usepackage{algorithmic}

\usepackage{newfloat}
\usepackage{listings}
\DeclareCaptionStyle{ruled}{labelfont=normalfont,labelsep=colon,strut=off} 
\floatstyle{ruled}
\newfloat{listing}{tb}{lst}{}
\floatname{listing}{Listing}
\usepackage{amsmath}
\usepackage{graphicx}
\usepackage{algorithm}
\usepackage{algorithmic}
\usepackage{multicol}
\usepackage{multirow}
\usepackage{bm}
\usepackage{graphicx}
\usepackage{booktabs}
\usepackage{cleveref}
\usepackage{bm}
\usepackage{amssymb}
\usepackage{tablefootnote}
\usepackage{threeparttable}

\title{Improving Multimodal Sentiment Analysis via Modality Optimization and Dynamic Primary Modality Selection}
\author{
    Dingkang Yang\textsuperscript{\rm 1,2},\, Mingcheng  Li\textsuperscript{\rm 1}$^{*}$,\, Xuecheng Wu\textsuperscript{\rm 3},\, Zhaoyu Chen\textsuperscript{\rm 1},\, Kaixun Jiang\textsuperscript{\rm 1}, \\
    Keliang Liu\textsuperscript{\rm 1}\, Peng Zhai\textsuperscript{\rm 1}\, Lihua Zhang\textsuperscript{\rm 1}\thanks{Corresponding authors.}
}
\affiliations{
    \textsuperscript{\rm 1}College of Intelligent Robotics and Advanced Manufacturing, Fudan University 
    \textsuperscript{\rm 2}ByteDance\\
    \textsuperscript{\rm 3}School of Computer Science and Technology, Xi'an Jiaotong University\\
     yangdingkang@bytedance.com, lihuazhang@fudan.edu.cn
}

\usepackage{bibentry}

\begin{document}

\maketitle

\begin{abstract}
Multimodal Sentiment Analysis (MSA) aims to predict sentiment from language, acoustic, and visual data in videos. However, imbalanced unimodal performance often leads to suboptimal fused representations. Existing approaches typically adopt fixed primary modality strategies to maximize dominant modality advantages, yet fail to adapt to dynamic variations in modality importance across different samples. Moreover, non-language modalities suffer from sequential redundancy and noise, degrading model performance when they serve as primary inputs. 
To address these issues, this paper proposes a \text{m}odality \text{o}ptimization and \text{d}ynamic primary modality \text{s}election framework (MODS). First, a Graph-based Dynamic Sequence Compressor (GDC) is constructed, which employs capsule networks and graph convolution to reduce sequential redundancy in acoustic/visual modalities. Then, we develop a sample-adaptive Primary Modality Selector (MSelector) for dynamic dominance determination. Finally, a Primary-modality-Centric Cross-Attention (PCCA) module is designed to enhance dominant modalities while facilitating cross-modal interaction. Extensive experiments on four benchmark datasets demonstrate that MODS outperforms state-of-the-art methods, achieving superior performance by effectively balancing modality contributions and eliminating redundant noise.
\end{abstract}


\section{Introduction}

In the modern digital era, individuals frequently share opinions and emotions through social media and e-commerce platforms. Sentiment analysis of such data holds broad application value ~\cite{gallagher2019application, tsai2021analyzing, drus2019sentiment,yang2022emotion,yang2023context}. Traditional unimodal sentiment analysis (USA) methods~\cite{ortis2019overview, xiao2021research} rely solely on a single data source (\textit{e.g.,} language or visual) and suffer from inherent limitations such as informational ambiguity and poor noise resistance. Emerging multimodal sentiment analysis (MSA) techniques enhance analytical accuracy by integrating multidimensional information from different modalities, better aligning with natural human emotional expression, where emotions are often conveyed through multiple channels, such as language content, facial expressions, and vocal tones.

In the MSA task~\cite{mai2022hybrid}, different modalities contribute unevenly to sentiment prediction. The language modality, due to its condensed emotional expression and high semantic density, is typically regarded as the primary information source, particularly excelling in opinion-oriented or conversational scenarios. Consequently, prior studies have developed a series of language-oriented MSA methods \cite{wu2021text,lei2023text,zhang2023learning} to use the advantages of language data. While these methods recognize intermodal disparities and improve performance, their static modality-dominant strategies have limitations: they capture population-level patterns but fail to adapt to dynamic sample-wise variations in modality dominance. 
Specifically, when non-language modalities dominate emotional expression in certain samples, forcibly prioritizing the language modality causes models to overlook critical affective cues from other modalities, thereby compromising prediction performance. 
Previous work~\cite{wang2023cross} identifies this issue and proposes the HCT-DMG for dynamic primary modality selection. However, HCT-DMG overlooks another key problem: the inherent sequential redundancy of non-language modalities. Compared to language modality, serialized representations of acoustic and visual modalities exhibit significantly lower information density, containing more repetitive and irrelevant features. Directly treating them as primary modalities may introduce noise interference, degrading fusion performance. Furthermore, asynchronous multimodal sequences restrict HCT-DMG to batch-level (rather than sample-level) primary modality selection, preventing true sample-wise adaptation.

To address these challenges, we propose a \text{m}odality \text{o}ptimization and \text{d}ynamic primary modality \text{s}election framework (MODS), a new MSA algorithm supporting sample-level dynamic modality selection. Specifically, we first design a Graph-based Dynamic Compressor (GDC) module to resolve sequential redundancy in non-language modalities via graph convolution operations. This module employs capsule networks \cite{sabour2017dynamic} for efficient graph structure modeling, compressing redundant information while eliminating noise in acoustic-visual modalities to enhance feature quality and emotional expressiveness. Then, we present a sample-adaptive MSelector module to transcend conventional unimodal dominance paradigms by dynamically determining the optimal primary modality for each input sample. Automatic modality selection based on sample characteristics enables flexible cross-scenario adaptability.  Finally, we introduce a Primary-modality-Centric Crossmodal Attention (PCCA) module to facilitate intermodal interaction and primary modality enhancement. These components improve MSA performance by simultaneously capturing dynamic modality dominance patterns and eliminating redundant interference, delivering a more robust and generalizable solution for complex multimodal tasks.
We develop the MODS and conduct comprehensive experiments on four public MSA datasets. Empirical results and analyses validate the framework’s effectiveness.

\section{Related Work}

\noindent \textbf{Ternary Symmetric-based Methods in MSA}.
Existing MSA methods primarily focus on multimodal representation learning and multimodal fusion~\cite{zadeh2018memory,hazarika2020misa,tsai2019multimodal,yang2022contextual,yang2022disentangled,yang2022learning,yang2023target}. Mainstream representation learning and fusion approaches typically treat different modalities equally. For instance, Zadeh \textit{et al.}~\cite{zadeh2018memory} proposed the Memory Fusion Network (MFN) that combines LSTM with attention mechanisms, utilizing their Delta-memory attention module to achieve cross-modal interaction at each timestep of aligned multimodal data. Hazarika \textit{et al.}~\cite{hazarika2020misa} introduced MISA, projecting each modality into two distinct subspaces to separately learn modality-invariant shared features and modality-specific characteristics. Tsai \textit{et al.}~\cite{tsai2019multimodal} created the MulT that employs pairwise cross-modal attention to align and fuse asynchronous multimodal sequences without manual alignment. Building upon MulT, Lv \textit{et al.}~\cite{lv2021progressive} proposed the PMR model, which extends bidirectional cross-modal attention to tri-directional attention involving all modalities while introducing a message hub to explore intrinsic inter-modal correlations for more efficient multimodal feature fusion. Liang \textit{et al.}~\cite{liang2021attention} developed the MICA architecture based on MulT, incorporating modality distribution alignment strategies into cross-modal attention operations to obtain reliable cross-modal representations for asynchronous multimodal sequences. Mai \textit{et al.}~\cite{mai2022hybrid} presented HyCon using hybrid contrastive learning for trimodal representation, employing three distinct contrastive learning models to capture inter-modal interactions and inter-class relationships after obtaining unimodal representations. 


\noindent \textbf{Language Center-based Methods in MSA}.
Another prominent line of research focuses on enhancing the weighting of language modality in MSA tasks to better leverage dominant modalities, aiming to overcome performance bottlenecks in MSA. For instance, existing efforts~\cite{delbrouck2020transformer,han2021bi,wang2023tetfn} employed transformer-based approaches to integrate complementary information from other modalities through language modality. Han \textit{et al}.~\cite{han2021improving} further improved this by maximizing mutual information between non-textual and textual modalities to filter out task-irrelevant modality-specific noise. Meanwhile, some works~\cite{li2022amoa,lin2022multimodal} introduced contrastive learning to capture shared features between non-language and language modalities. Furthermore, Wu \textit{et al.}~\cite{wu2021text} proposed a Text-Centric Shared-Private (TCSP) framework that distinguishes between shared and private semantics of textual and non-textual modalities, effectively fusing text features with two types of non-text features. Building upon Transformer~\cite{vaswani2017attention} architectures, impressive methods~\cite{lei2023text,zhang2023learning, wu2023denoising} have been designed that establish language modality's dominant role to achieve sequence fusion. 
Although the aforementioned types of MSA methods have achieved encouraging results, performance bottlenecks still exist. Approaches that treat each modality equally may cause the model to be distracted by secondary modalities, while methods focused on the language modality lack flexibility and struggle to adapt to variations in dominant modality across samples. 

\section{Methodology}

\begin{figure*}[!h]
\centering 
\includegraphics[width=1\linewidth]{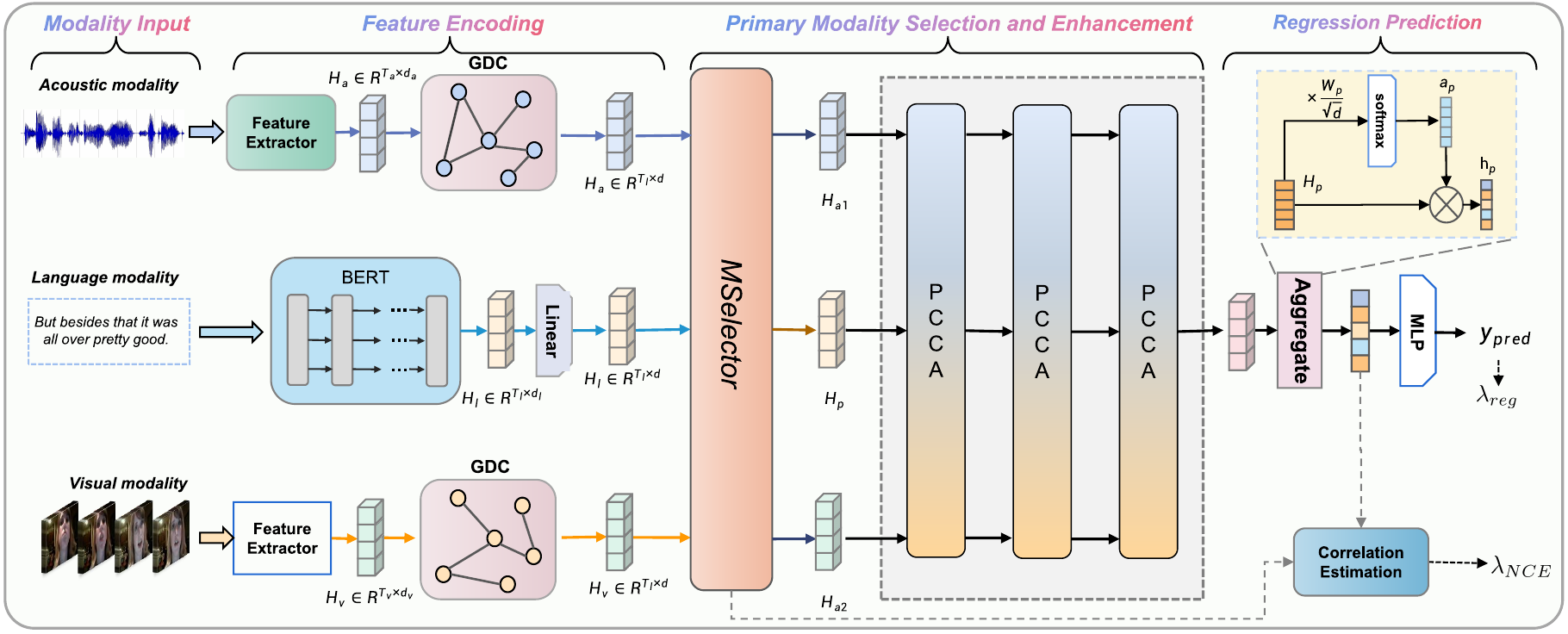}
\caption{The overall architecture of the proposed MODS framework.} 
\label{mods}
\end{figure*}

\subsection{Problem Statement}
The MSA task focuses on detecting sentiments by leveraging the language (\textit{l}), visual (\textit{v}), and acoustic (\textit{a}) modalities from the video clips. Given unimodal sequences from these modalities as $ X_{m} \in \mathbb{R}^{T_{m} \times d_{m}} $, where $m \in \left\{ {l, v, a} \right\}$. $ T_{m} $ and $ d_{m} $ are used to represent the sequence length and the feature dimension of the modality \textit{m}, respectively. 

\subsection{Framework Overview}
The overall architecture of the MODS algorithm is illustrated in Figure~\ref{mods}. First, preliminary feature extraction is performed on each modality using corresponding methods to obtain unimodal sequence features \( H_m  \), where \( m \in \{ l, a, v \} \). For the language features, after extraction via BERT, a linear layer projects the feature dimension to \( d \), yielding \( H_l \in \mathbb{R}^{T_{l} \times d} \). Meanwhile, the acoustic and visual features are compressed to the same dimension as the language features through a graph convolutional sequence compression module, resulting in \( H_a \in \mathbb{R}^{T_{l} \times d} \) and \( H_v \in \mathbb{R}^{T_{l} \times d} \), respectively.  
The three unimodal features are then fed into the Primary Modality Selector (MSelector) to determine the primary modality \( p \) and secondary modalities \( a1 \), \( a2 \) along with their weighted sequence feature outputs \( H_p \), \( H_{a1} \), and \( H_{a2} \).  
After primary modality selection, the three modality features are progressively processed by the Primary-modality-Centric Crossmodal Attention (PCCA) module for multimodal interaction and fusion. The final output is the enhanced fused feature \( H_p \).  
Ultimately, \( H_p \) undergoes adaptive aggregation to produce a one-dimensional vector \( h_p \), which is then passed through a multilayer perceptron (MLP) to obtain the sentiment prediction result \( y_{pred} \).

\subsection{Graph-based Dynamic Compression}
RNNs \cite{hochreiter1997long} were widely used to study sequential semantics in feature embedding modeling. However, it suffers from slow training, difficulty in capturing long-range dependencies due to their recurrent nature, and gradient-related issues. Transformers~\cite{vaswani2017attention} enable parallel computation and improve long-range dependency modeling, but rely on attention-weighted summation, making explicit temporal relationship modeling challenging. Both lack sequence compression capability: RNNs process data step-by-step inefficiently, while Transformers incur redundant computations and suboptimal weight allocation. In contrast, graph structures enhance information efficiency via direct node connections, and graph convolutional networks (GCNs)~\cite{kipf2016semi} improve stability through neighbourhood aggregation, mitigating gradient problems. Thus, we propose the Graph-based Dynamic Compression (GDC) module to project non-language sequences into graph space, exploring intra-modal dependencies while compressing redundant long sequences. 

\begin{figure}[t]
\centering 
\includegraphics[width=1\linewidth]{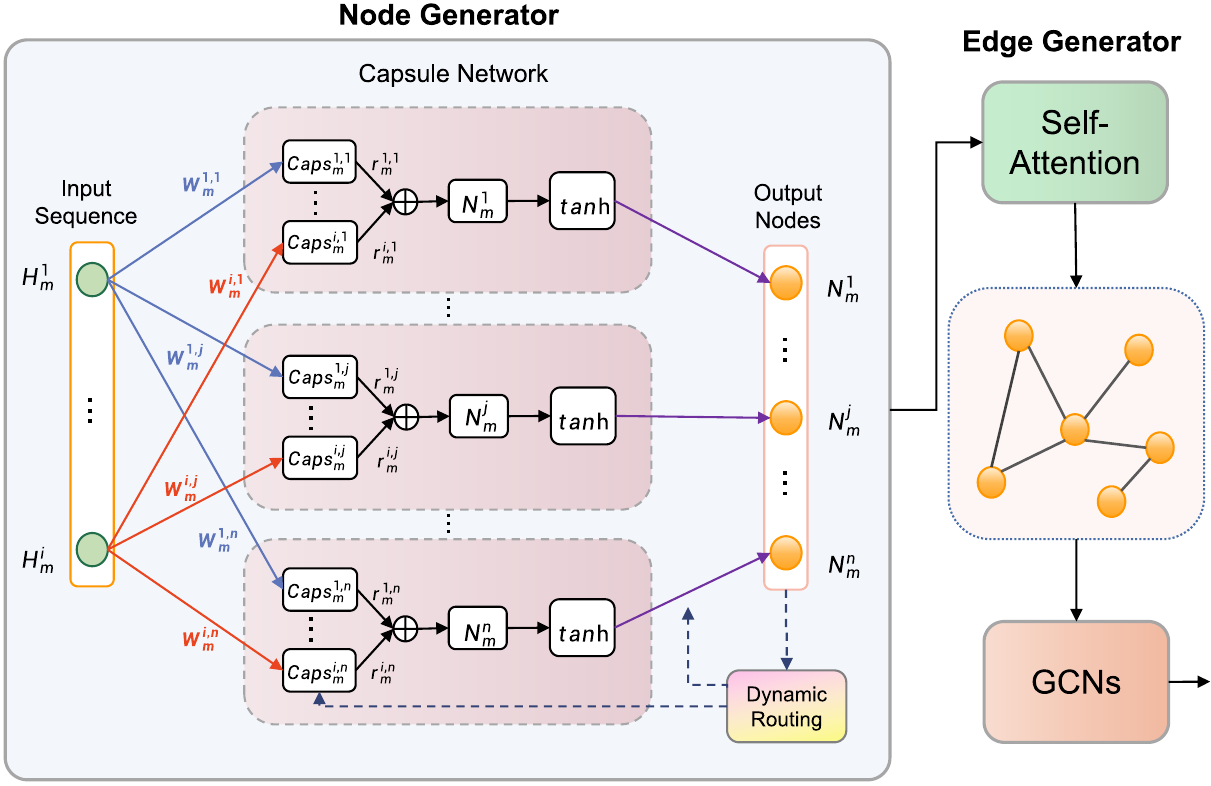}
\caption{The  architecture of the proposed GDC module.} 
\label{gdc}
\end{figure}

In prior sequence modeling frameworks, graph node construction generally follows one of two strategies: treating features at each timestep as individual nodes, which preserves sequence fidelity at the cost of computational redundancy, or employing temporal slice pooling, which improves efficiency but may compromise fine-grained temporal information. Instead of these, we propose a capsule network-based solution, where its dynamic routing mechanism adaptively learns node importance weights, and vectorized feature representation preserves directional and semantic information. This enables sequence compression while avoiding the loss of critical details. The detail of the GDC module is shown in Figure~\ref{gdc}.
We first compress the sequence information into an appropriate number of nodes using a capsule network. To facilitate model training and subsequent cross-modal interactions, we align the compressed sequence lengths of acoustic and visual modalities with that of the language modality. Specifically, for the sequence features $H_m$ of modality $m \in{a,v}$, the capsule $Caps$ construction process is defined as follows:
\begin{equation}
   Caps^{i,j}_{m} = W_{m}^{ij} H_{m}^{i},
   \label{cap_generate}
\end{equation}
where $H_{m}^{i} \in R^{d_m}$denotes the feature at the $i$-th timestep of sequence $H_m$, $ W_{m}^{ij} \in R^{d  \times d_m}$ represents the trainable parameters, and $ Caps^{i,j}_{m} \in R^d$ is the capsule created from the $i$-th timestep feature for constructing the $j$-th node.
Subsequently, the graph node representations are generated based on the defined capsules and the dynamic routing algorithm:
\begin{equation}
    N_{m}^{j} = \sum_{i} Caps^{i,j}_{m} \times r_{m}^{i,j},
    \label{node_generate}
\end{equation}
where $r_{m}^{i,j}$ is the is the routing coefficient, which is updated through an iterative optimization process:
\begin{equation}
    r_{m}^{i,j} = \frac{\exp(b_{m}^{i,j})}{\sum_j \exp(b_{m}^{i,j})},
    \label{r_generate}
\end{equation}
\begin{equation}
    b_{m}^{i,j} \leftarrow b_{m}^{i,j} + Caps^{m}_{i,j} \odot tanh( N_{m}^{j}).
    \label{r_update}
\end{equation}
Here, $b_{m}^{i,j}$ is the unnormalized routing coefficient initialized to 0. After normalization, the initial value of $r_{m}^{i,j}$ is $\frac1n$ (where $n$ is the number of capsules), meaning each capsule is initially assigned equal routing coefficients. According to the properties of capsule networks, during the iterative process of dynamic routing, capsules receiving noise or redundant information will be gradually assigned smaller weights, while capsules containing important affective information will obtain larger weights. Therefore, using capsule networks to create graph nodes helps improve the quality of acoustic-visual features and increase the density of effective information in sequences.
After obtaining the node representations of the modality feature graph, we employ a self-attention mechanism to explore the edge weight relationships between nodes in the feature graph:
\begin{equation}
    E_{m} = ReLU\left(\frac{\left(W_{m}^{q} N_{m}\right)^{T} \left(W_{m}^{k} N_{m}\right)}{\sqrt{d}}\right).
\end{equation}
The edge construction process, empowered by the self-attention mechanism, focuses on the most relevant information in the modality feature graph to generate edge representations. This process effectively strengthens the dependencies between nodes, ensuring that the generated edge representations can more accurately reflect the inter-node relationships within the graph.

GCNs are pivotal for graph-based learning. GCNs enable simultaneous processing of complex data, capture essential global information, and demonstrate high efficacy in learning both nodes and edges. Therefore, for the modality feature graph $G_m=(N_m, E_m)$, we employ GCN to perform representation learning on the graph. The specific formulation is as follows:

\begin{equation}
H_{m}^{l} = ReLU\left( D^{-\frac{1}{2}}_{m} E_{m} D^{-\frac{1}{2}}_{m} H_{m}^{l-1} W_{m}^{l} + b_{m}^{l} \right),
\end{equation}
where $H_{m}^{l}$ is the output of $l$-th GCN, $m \in \left\{ l, a, v \right\}$,$D_{m}$ is the degree matrix of $E_{m}$, $W_{m}^{l}$ and $b_{m}^{l}$ are the learnable parameters of the $l$-th GCN layer. The initial input to the GCN is the set of node representations from the modality feature graph, denoted as $H_{m}^{0} = N_{m}^{j}$.

\subsection{Primary Modality Selection}
The adaptive dynamic primary modality selection module (MSelector) aims to enable the network to autonomously select the primary modality through training, eliminating the need for manual intervention. The core mechanism of this module involves dynamically evaluating the importance of each modality by learning its trainable parameters.
First, it performs adaptive aggregation on each modality's sequential features, transforming them into one-dimensional feature vectors. Specifically, for $H_{m}\in R^{T_{m} \times d}$, the attention weight matrix is first computed as follows:

\begin{equation}
    a_{m} = softmax\left( \frac{H_{m}W_{m}}{\sqrt{d}}\right)^{T}\in R^{1 \times T_{m}},
\end{equation}
where $m \in \left\{ l, a, v \right\}$. $W_{m} \in R^{ d}$ is the linear projection parameters and $a_{m}$ represents the attention weight matrix for $H_{m}$. Subsequently, the attention weights are multiplied with the sequential features to obtain the aggregated vector:
\begin{equation}
h_{m} = a_{m}H_{m} \in R^{1\times d}.
\end{equation}
Subsequently, the vectors from the three modalities are concatenated and fed into a Multilayer Perceptron (MLP) for processing:
\begin{equation}
   w=softmax(MLP( concat(h_a, h_l, h_v))).
\end{equation}
The output of the MLP is a three-dimensional vector, which is processed through a softmax function to generate three weight values that sum to 1:
\begin{equation}
    w = [w_a, w_t, w_v], \quad w_a + w_t + w_v = 1.
\end{equation}
Each weight value represents the importance degree of a corresponding modality, where a higher weight indicates that the model considers the modality to have greater contribution. The modality with the highest weight is identified as the primary modality $p$:
\begin{equation}
    p = \arg\max(w_a, w_t, w_v).
\end{equation}
To further enhance the influence of the primary modality while distinguishing between the suboptimal modality $a1$ and $a2$, we multiply these weight values with their corresponding modality features, and use the resulting products as inputs to the subsequent network:
\begin{equation}
    H_{a_1} = w_a \cdot H_a, \quad H_{a_2} = w_t \cdot H_t, \quad H_p= w_v \cdot H_v.
\end{equation}
It should be noted that the formula here is written with modality $v$ as the primary modality for illustration purposes. During training and inference, the primary modality dynamically varies based on the model's selection.

\subsection{Primary-modality-Centric Interaction}
\begin{figure}[t]
\centering 
\includegraphics[width=1\linewidth]{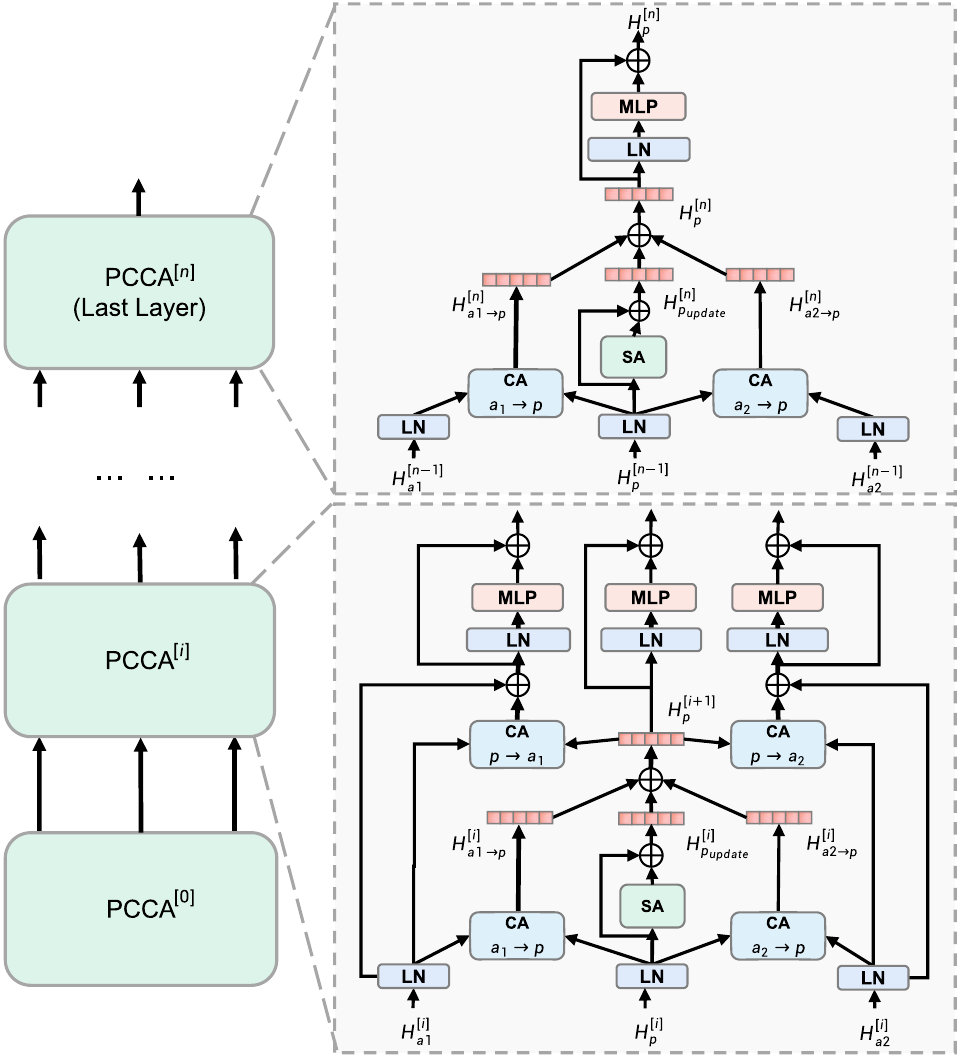}
\caption{The  architecture of the proposed PCCA module.} 
\label{pcca}
\end{figure}

To capture element correlations among multimodal sequences and enhance the primary modality, we design the PCCA module to enable cross-modal interactions. In the PCCA module, a two-step process is employed to facilitate information flow among three modalities. Initially, information from the suboptimal modality flows to the primary modality via the ${CA}_{a\rightarrow p}^{\lbrack i\rbrack}$ block and is then fused. Subsequently, the fused information flows from the primary modality to the suboptimal modality via the ${CA}_{p\rightarrow a}^{\lbrack i\rbrack}$ block. This module ensures a mutual information flow and progressive enhancement across all modalities.  The architecture of $PCCA^{\lbrack i\rbrack}$ is shown in Figure~\ref{pcca}, and the superscript $\lbrack i\rbrack$ indicates the $i$-th modality reinforcement processes. Its inputs are $H_{p}^{\lbrack i\rbrack}$, $H_{a1}^{\lbrack i\rbrack}$,  and $H_{a2}^{\lbrack i\rbrack}$  while its outputs are the reinforced features of these three modalities as $H_{p}^{\lbrack i+1\rbrack}$, $H_{a1}^{\lbrack i+1\rbrack}$ and $H_{a2}^{\lbrack i+1\rbrack}$:
\begin{equation}
       \begin{aligned}
           H_{p}^{\lbrack i+1\rbrack},  H_{a1}^{\lbrack i+1\rbrack},  H_{a2}^{\lbrack i+1\rbrack} &= ~{PCCA}^{\lbrack                i\rbrack}\left( {H_{p}^{\lbrack i\rbrack},H_{a1}^{\lbrack i\rbrack},H_{a2}^{\lbrack i\rbrack}} \right).
       \end{aligned}
\end{equation}
Specifically, we first perform a layer normalization ($LN$) on the features  $
H_{m}^{\lbrack i\rbrack} $,  where $H_{m}^{\lbrack 0\rbrack} = H_{m}$ and $m \in \left\{ {p,a1,a2} \right\}$. Then, two cross-attention and one self-attention are simultaneously implemented. The cross-attention block ${CA}_{a\rightarrow l}^{\lbrack i\rbrack}$ takes $H_{p}^{\lbrack i\rbrack}$ as  Quary and  $H_{a}^{\lbrack i\rbrack}$ as Key and Value, and obtain $H_{a\rightarrow p}^{\lbrack i\rbrack}$ with $a \in \left\{ {a1,a2} \right\}$ :
\begin{equation}
       \begin{aligned}
           H_{a\rightarrow p}^{\lbrack i\rbrack} &= ~{CA}_{a\rightarrow p}^{\lbrack                i\rbrack}\left( {H_{a}^{\lbrack i\rbrack},H_{p}^{\lbrack  i\rbrack}} \right).
       \end{aligned}
\end{equation}
Meanwhile, the self-attention block ${SA}_{p}^{\lbrack i\rbrack}$ takes $
H_{p}^{\lbrack i\rbrack}$ as input to obtain $H_{p_{update}}^{\lbrack{i + 1}\rbrack}$:
\begin{equation}
    \begin{aligned}
        H_{p_{update}}^{\lbrack{i}\rbrack} &= ~{SA}_{p}^{\lbrack i\rbrack}\left( H_{p}^{\lbrack i\rbrack} \right)+ H_{p}^{\lbrack i\rbrack}.
    \end{aligned}
\end{equation}
Here, ${SA}_{p}^{\lbrack i\rbrack}$ aims to capture the contextual relationships inside the modality and realize the modality's self-reinforcement. Now we obtain primary modality features that capture both intra-modal and inter-modal  dependencies, and then we add them together to get a complete enhancement of the primary modality:
\begin{equation}
    \begin{aligned}
        H_{p}^{\lbrack{i+1}\rbrack} &= ~H_{p_{update}}^{\lbrack i\rbrack}+ \sum\limits_{a \in \{ a1,a2\}}H_{a\rightarrow p}^{\lbrack i\rbrack}.
    \end{aligned}
\end{equation}
Since $H_{p}^{\lbrack{i+1}\rbrack}$ contains information from all modalities, we then perform two cross-attention operations ${CA}_{p\rightarrow a}^{\lbrack i\rbrack}$ to pass the information of the enhanced language modality to the visual/acoustic modalities,  which is expressed as follows:
\begin{equation}
       \begin{aligned}
           H_{p\rightarrow a}^{\lbrack i\rbrack} &= ~{CA}_{p\rightarrow a}^{\lbrack                i\rbrack}\left( {H_{p}^{\lbrack{i+1}\rbrack},H_{a}^{\lbrack                i\rbrack}} \right).
       \end{aligned}
\end{equation}
In doing so, we utilize the primary modality as a bridge to implicitly realize the information flow among all three modalities.
Subsequently, we process $H_{p}^{\lbrack{i + 1}\rbrack}$ and $H_{p\rightarrow a}^{\lbrack i\rbrack}$ by a Position-wise Feed-Forward layer ($PFF$) with skip connection:
\begin{equation}
    H_{a}^{\lbrack i+1\rbrack} = PFF\left( {LN\left( H_{p\rightarrow a}^{\lbrack i\rbrack} \right)} \right) + H_{p\rightarrow a}^{\lbrack i\rbrack},
\end{equation}
\begin{equation}
    H_{p}^{\lbrack{i + 1}\rbrack} = PFF\left( {LN\left( H_{p}^{\lbrack{i + 1}\rbrack} \right)} \right) + H_{p}^{\lbrack{i + 1}\rbrack}.
\end{equation}
It should be noted that the final PCCA module retains only the ${CA}_{a\rightarrow p}$ operation, with its enhanced output feature $H_p$ being used for downstream regression tasks. 
Since direct interactions between suboptimal modalities make it easy to generate interference information, our proposed PCCA module uses the primary modality as a bridge so that suboptimal modalities' information can be fused under the primary modality's feature monitoring, thereby reducing the generation of unnecessary semantics. 

\begin{table*}[t]
\centering
\renewcommand{\arraystretch}{1.0}
\setlength{\tabcolsep}{3pt}
\resizebox{\textwidth}{!}{%
\begin{tabular}{lccccc|ccccc}
\toprule
\multirow{2}{*}{Methods }& \multicolumn{5}{c|}{MOSI Dataset}        & \multicolumn{5}{c|}{MOSEI Dataset}       \\ \cmidrule{2-11} 
                         & $MAE \downarrow$ & $Corr \uparrow$ & $Acc_7 \uparrow$ & $Acc_2  \uparrow$ & $F1  \uparrow$ & $MAE \downarrow$ & $Corr \uparrow$ & $Acc_7 \uparrow$ & $Acc_2  \uparrow$ & $F1  \uparrow$ \\  
\midrule
TFN~\cite{zadeh2017tensor}$^\dagger$       & 0.947& 0.673& 34.46& 77.99/79.08& 77.95/79.11& 0.572& 0.714& 51.60& 78.50/81.89& 78.96/81.74\\[1 ex]
LMF~\cite{liu2018efficient}$^\dagger$       & 0.950& 0.651& 33.82& 77.90/79.18& 77.80/79.15& 0.576& 0.717& 51.59& 80.54/83.48& 80.94/83.36\\[1 ex]
ICCN~\cite{sun2020learning}& 0.862 & 0.714 & 39.0  & -/83.0      & -/83.0      & 0.565 & 0.713 & 51.6  & -/84.2      & -/84.2 \\[1 ex]
MulT~\cite{tsai2019multimodal}$^\dagger$& 0.879& 0.702& 36.91& 79.71/80.98& 79.63/80.95& 0.559& 0.733& 52.84& 81.15/84.63& 81.56/84.52\\[1 ex]
MISA~\cite{hazarika2020misa}$^\dagger$& 0.776& 0.778& 41.37& 81.84/83.54& 81.82/83.58& 0.568 & 0.724 & -     & 82.59/84.23 & 82.67/83.97 \\[1 ex]
Self-MM~\cite{yu2021learning}$^\dagger$& \text{0.708}& 0.796& 46.67& 83.44/85.46& \text{83.36}/85.43& 0.531& 0.764& 53.87& \text{83.76}/85.15& \text{83.82}/84.90\\[1 ex]
MMIM~\cite{han2021improving}$^\ast$         & 0.718 & \text{0.797}& 46.64 & 83.38/\text{85.82}& 83.29/\text{85.81}& 0.537 & 0.759 & 53.42 & 82.08/85.14 & 82.51/85.11 \\[1 ex]
MSG ~\cite{lin2023dynamically}               & 0.748 & 0.782 & 47.3& -/85.6 & /85.6 & 0.583 & 0.787 & 52.8 & -/85.4 & -/85.4 \\[1 ex]
PriSA ~\cite{ma2023multimodal}$^\ast$            & 0.719 & 0.782 & 47.1& 83.3/85.4& 83.1/85.4 & 0.536& \text{0.761}& 53.95& 82.1/85.2& 83.3/\text{85.2}\\[1 ex]
DMD \cite{li2023decoupled}$^\ast$                                           & -              & -              & 43.9  & - /84.9          & - /85.0          & -              & -              & 53.1          & - /85.2          & - /85.2         \\[1 ex]
MIM \cite{zeng2023multimodal}$^\ast$                                           & 0.718          & 0.792          & 46.4  & - /84.8          & - /84.8          & 0.579          & 0.779          & 51.8          & - /85.7          & - /85.6         \\[1 ex]
DTN \cite{zeng2024disentanglement}$^\ast$                                           & 0.716          & 0.790          & \text{47.5}  & - /85.1          & - /85.1          & 0.572          & 0.765          & 52.3          & - /85.5          & - /85.5         \\ [1 ex]
\textbf{MODS (Ours)} & \textbf{0.688} & \textbf{0.798} & \textbf{49.27} & \textbf{83.53}/\textbf{85.83} & \textbf{83.75}/\textbf{85.96} &  \textbf{0.527}& \textbf{0.772} & \textbf{54.32}& \textbf{84.52}/\textbf{85.88}& \textbf{84.5/86.14}\\
\bottomrule
\end{tabular}
}
\caption{Comparison results on the MOSI and MOSEI datasets.  $\dagger$: the results from \cite{mao2022m}; $\ast$: the results are reproduced from the open-source codebase with hyper-parameters provided in original papers.
For $Acc_2$ and $F1$, we have two sets of non-negative/negative (left) and positive/negative (right) evaluation results. Bold represents the best results, respectively.}
\label{tab_mosi_mosei}
\end{table*}

\begin{table*}[t]
\centering
\renewcommand{\arraystretch}{1.0}
\setlength{\tabcolsep}{6pt}
\resizebox{\textwidth}{!}{%
\begin{tabular}{lcccccc|cccccc}
\toprule
\multirow{2}{*}{Methods}& \multicolumn{6}{c|}{SIMS Dataset}        & \multicolumn{6}{c|}{SIMSv2 Dataset}       \\ \cmidrule{2-13} 
                         & $MAE \downarrow$ & $Corr \uparrow$ & $Acc_2 \uparrow$ & $Acc_3  \uparrow$ & $Acc_5  \uparrow $ & $F1  \uparrow$ & $MAE \downarrow$ & $Corr \uparrow$ & $Acc_2 \uparrow$ & $Acc_3  \uparrow$ & $Acc_5  \uparrow $ & $F1  \uparrow$ \\  
\midrule
TFN~\cite{zadeh2017tensor}$^\dagger$&  0.432& 0.591 & 78.3  & 65.12 &39.3  & 78.62 & 0.329 & 0.640 & 77.95 & 70.21 & 51.93 & 77.74  \\[1 ex]
LMF~\cite{liu2018efficient}$^\dagger$&  0.441& 0.576& 77.77& 64.68&40.53&77.88 &  0.367& 0.557& 74.18& 64.90&47.79&73.88\\[1 ex]
MulT~\cite{tsai2019multimodal}$\ast$&  0.453& 0.564& 78.56& 64.77&37.94& 79.66  &  \text{0.304}& \text{0.705}& 79.3& 72.63&\text{53.29}& 79.43\\[1 ex]
Self-MM~\cite{yu2021learning}$^\dagger$& 0.425& \text{0.595}& \text{80.04}& 65.47&41.53& \text{80.44} & 0.322& 0.678& 79.11& \text{72.34}&53.0& 79.05 \\[1 ex]
CENet~\cite{wang2022cross}$^\dagger$& 0.471 & 0.534 & 77.90 & 62.58 & 33.92  & 77.53 & 0.310 & 0.699 & 79.56    & 73.10   & 53.04    &\textbf{79.63}\\[1 ex]
ALMT~\cite{zhang2023learning}$^\ast$         & \text{0.408}& 0.594& 78.77& \text{65.86}&\text{43.11}&78.71 & 0.308& 0.700& \text{79.59}& 71.86&52.90&79.51\\[1 ex]
DMD \cite{li2023decoupled}$^\ast$                                           & 0.412              & 0.586              & 78.33  &  65.23         & 44.26         & 79.21     & 0.305        &  0.702             &  78.87         & 72.01          & 53.18   & 79.21     \\[1 ex]
MIM \cite{zeng2023multimodal}$^\ast$                                           & 0.420          & 0.592         & 78.98  & 65.12          & 44.98          &  78.70         &  0.310          &  0.694       &  77.56        & 71.45   & 52.87  & 78.56         \\[1 ex]
DTN \cite{zeng2024disentanglement}$^\ast$                                         &           0.419&          0.593 &   79.45&           65.67&           44.26&           79.47&           0.302&       0.701& 78.29&               72.56 &     53.71 &  78.12        \\[1 ex] 
\textbf{MODS (Ours)} & \textbf{0.407} & \textbf{0.605} & \textbf{80.96} & \textbf{66.74} & \textbf{45.51} &  \textbf{80.94} 
                & \textbf{0.297} & \textbf{0.712}& \textbf{79.59} & \textbf{73.69} & \textbf{55.51} & \text{79.53}\\
\bottomrule
\end{tabular}
}
\caption{Comparison results on the SIMS and SIMSv2 datasets. $\dagger$: the results from \cite{mao2022m}; $\ast$: the results are reproduced from the open-source codebase with hyper-parameters provided in original papers. Bold represents the best results, respectively.}
\label{tab_sims_simsv2}
\end{table*}

\subsection{Training Objective Optimization}
In the early stages of training, the selection of the primary modality may be influenced by weight initialization or data distribution, leading to instability in the choice. To address this, we introduce the InfoNCE loss to enhance the stability of primary modality selection and constrain the model to retain key information across different modalities. Specifically, for the output $H_p$ of the PCCA module, we first aggregate it into a one-dimensional vector $h_p$ using the adaptive aggregation method described above. Then, we construct a reverse prediction path $\mathcal{F}$ from the feature $h_{p}$ to the unimodal feature $h_{m}$, and measure the correlation between them using the following normalized similarity function:
\begin{equation}
    \text{sim}(h_m, h_{p}) = \exp\left(\frac{h_m}{\|h_m\|^2} \odot \frac{\mathcal{F}(h_p)}{\|\mathcal{F}(h_p)\|^2}\right),
\end{equation}
where $\mathcal{F}$ takes $h_{p}$ as input and produces a prediction for $h_{m}$, $\| \cdot \|^2$ denotes the Euclidean norm normalization, and $\odot$ represents element-wise multiplication. Then, the above similarity function is incorporated into the noise-contrastive estimation framework to generate the InfoNCE loss:
\begin{equation}
    \mathcal{L}_{\text{NCE}}^{f,m} = -\mathbb{E}_{h_m,h_p} \left[ \log \frac{\exp \left( \text{sim}(h_m^+, \mathcal{F}(h_p)) \right)}{\sum_{k=1}^{K} \exp \left( \text{sim}(\tilde{h}_m^k, \mathcal{F}(h_p)) \right)} \right].
\end{equation}
$\tilde{h}_m = \{\tilde{h}^1,\tilde{h}^2...\tilde{h}^K\}$ represents all other samples in the same batch during training, where $K$ is the batch size. These samples are treated as negative samples for contrastive learning. Finally, the InfoNCE loss for all modalities is computed as follows:
\begin{equation}
    \mathcal{L}_{\text{NCE}} =   \mathcal{L}_{\text{NCE}}^{p,l
    } + \mathcal{L}_{\text{NCE}}^{p,a} + \mathcal{L}_{\text{NCE}}^{p,v}. 
\end{equation}
In the output, we input the representation $h_p$ into an MLP to predict sentiment score $y_{pred}$. Given the predictions $y_{pred}$ and the ground truth $y_{true}$, we calculate the task loss $ \mathcal{L}_{reg}$ by mean absolute error (MAE). Finally, we training MODS by the union loss $\mathcal{L}_{task}$:
\begin{equation}
    y_{pred} = MLP\left( h_{p};\theta_{\Phi} \right),
\end{equation}
\begin{equation}
   \mathcal{L}_{reg} = \frac{1}{N} \sum_{i=1}^{N} \lvert y_{true} - y_{pred} \rvert,
\end{equation}
\begin{equation}
    \mathcal{L}_{task} = \mathcal{L}_{reg} + \alpha \mathcal{L}_{\text{NCE}},
\end{equation}
where $\alpha$ is a parameter that balances the loss contribution.

\section{Experiments}

\subsection{Datasets and Evaluation Metrics}
We conduct experiments on four publicly benchmark datasets of MSA, including MOSI~\cite{zadeh2016mosi}, 
MOSEI \cite{zadeh2018multimodal},  SIMS \cite{yu2020ch}, and SIMSv2 \cite{liu2022make}. In contrast to the MOSI and MOSEI, both SIMS and SIMSv2 prioritize balanced modality-specific sentiment dominance, avoiding a clear trend where any single modality dominates emotional expression. This design better validates the effectiveness of MODS.
We use the accuracy of 3-class ($Acc_3$) and 5-class ($Acc_5$) on SIMS and SIMSv2, the accuracy of 7-class ($Acc_7$) on MOSI and MOSEI, and the accuracy of 2-class ($Acc_2$), Mean Absolute Error ($MAE$), Pearson Correlation ($Corr$), and F1-score ($F1$) on all datasets. In particular, higher values indicate better performance for all metrics except $MAE$.

\subsection{Implementation Details}
Following the \cite{yu2021learning}, we use unaligned raw data in all experiments. All models are built on the Pytorch toolbox \cite{paszke2019pytorch} with two Quadro RTX 8000 GPUs. The Adam optimizer~\cite{kingma2014adam} is adopted for network optimization. The training duration of each model is governed by an early-stopping strategy with the patience of 25 epochs for MOSI, SIMS and SIMSv2 and 15 epochs for MOSEI. 
For MOSI, MOSEI, SIMS, and SIMSv2, the detailed hyper-parameter settings are as follows: the learning rates are $\{3e-5, 1e-5, 1e-5,1e-5\}$, the batch sizes are $\{32, 64, 32, 32\}$, the hidden size are $\{128, 128, 64, 128\}$, the PCCA layers are $\{3, 3, 3, 4\}$, the coefficient $\alpha$ are $\{0.1, 0.1, 0.01, 0.01\}$, and weight decay are $\{1e-3, 1e-3, 1e-2, 1e-2\}$.
The hyper-parameters are determined based on the validation set.

\subsection{Comparison with State-of-the-art Methods}
Tables \ref{tab_mosi_mosei} and \ref{tab_sims_simsv2} present comparison experimental results of different methods on the MOSI, MOSEI, SIMS, and SIMSv2 datasets. It can be observed that MODS achieves the best performance on most metrics across all datasets, surpassing previous methods to become the new state-of-the-art (SOTA).  Compared to methods that treat all modalities equally or fix a primary one, MODS dynamically selects the primary modality to reduce interference from heterogeneous emotions. It exhibits lower error rates and stronger fine-grained emotion discrimination. This observation also validates the rationality and effectiveness of the sample-level dynamic primary modality selection approach.

\begin{table}[t]
\setlength{\tabcolsep}{2pt}
\centering
\small
\resizebox{\columnwidth}{!}{%
\begin{tabular}{lcccccccc}
\toprule
\multirow{2}{*}{Setting} &\multicolumn{3}{c}{SIMS Dataset}	& \multicolumn{3}{c}{MOSI Dataset} \\
\cmidrule{2-7} 
	& $MAE \downarrow$  & $Corr  \uparrow$ & $Acc_5  \uparrow$ &$MAE \downarrow$  & $Corr  \uparrow$ & $Acc_7  \uparrow$\\ 
\midrule
\textbf{MODS (Full)} & \textbf{0.407} & \textbf{0.605} & \textbf{45.51} & \textbf{0.688} & \textbf{0.798} & \textbf{49.27} \\ 
\midrule 
w/o GDC & 0.428 & 0.584 & 42.01 &  0.731 & 0.792 & 45.34\\
w/o Caps & 0.426 & 0.582& 43.33 & 0.727 & 0.787  & 48.27\\
\midrule
$l$-oriented & 0.417 & 0.595 & 43.33 & 0.709 & 0.789 & 45.92\\
$a$-oriented & 0.419 & 0.599 & 43.76 & 0.717 & 0.782& 45.34\\
$v$-oriented & 0.418 & 0.586 & 42.45  & 0.713 & 0.785& 46.94\\
\midrule
w/o PCCA & 0.444 & 0.554 & 42.89 &  0.738 & 0.775 & 45.48\\
\bottomrule
\end{tabular}
}
\caption{Ablation study results of MODS's components on SIMS and MOSI datasets.}
\label{tab_mods_module_ablation}
\end{table}

\subsection{Ablation Study}
To further verify the effectiveness of the proposed method, we launched elaborate ablation experiments on both SIMS and MOSI datasets, as shown in \Cref{tab_mods_module_ablation}.

\noindent \textbf{Effectiveness of GDC Module.} For the GDC Module, we conduct two comparisons: (1) experiments without GDC, and (2) experiments excluding the capsule network during graph construction. The results show that two ablation approaches lead to significant performance degradation, with the complete removal of the GDC module causing more severe performance deterioration. This occurs because under unbalanced modality information density, the selection of primary modalities and cross-modal interactions are both adversely affected, making it impossible to fully utilize the effective information from non-language modalities. These findings demonstrate the importance of performing sequence compression on non-language modalities and highlight the crucial role of using capsule networks for graph node construction. 

\noindent \textbf{Importance of MSelector Module.} To validate the importance of the  MSelector module, we compare MODS with models that fixed single modality as the primary modality, respectively. The experimental results demonstrate that fixing the primary modality leads to performance degradation, indicating that dynamic primary modality selection outperforms static modality assignment.

\noindent \textbf{Effect of PCCA Module.} To validate the proposed PCCA module, we compare the model's performance with and without PCCA. Specifically, after primary modality selection, we directly perform sentiment regression using only the selected primary modality, bypassing inter-modal interaction. The consistent performance drop across metrics demonstrates that even with optimal primary modality selection, ignoring complementary relationships between modalities severely limits prediction accuracy. This confirms that exhaustive inter-modal interaction, where auxiliary modalities supplement the primary modality with contextual information, is indispensable for achieving robust predictions.

\begin{figure}[t]
\centering 
\includegraphics[width=1\linewidth]{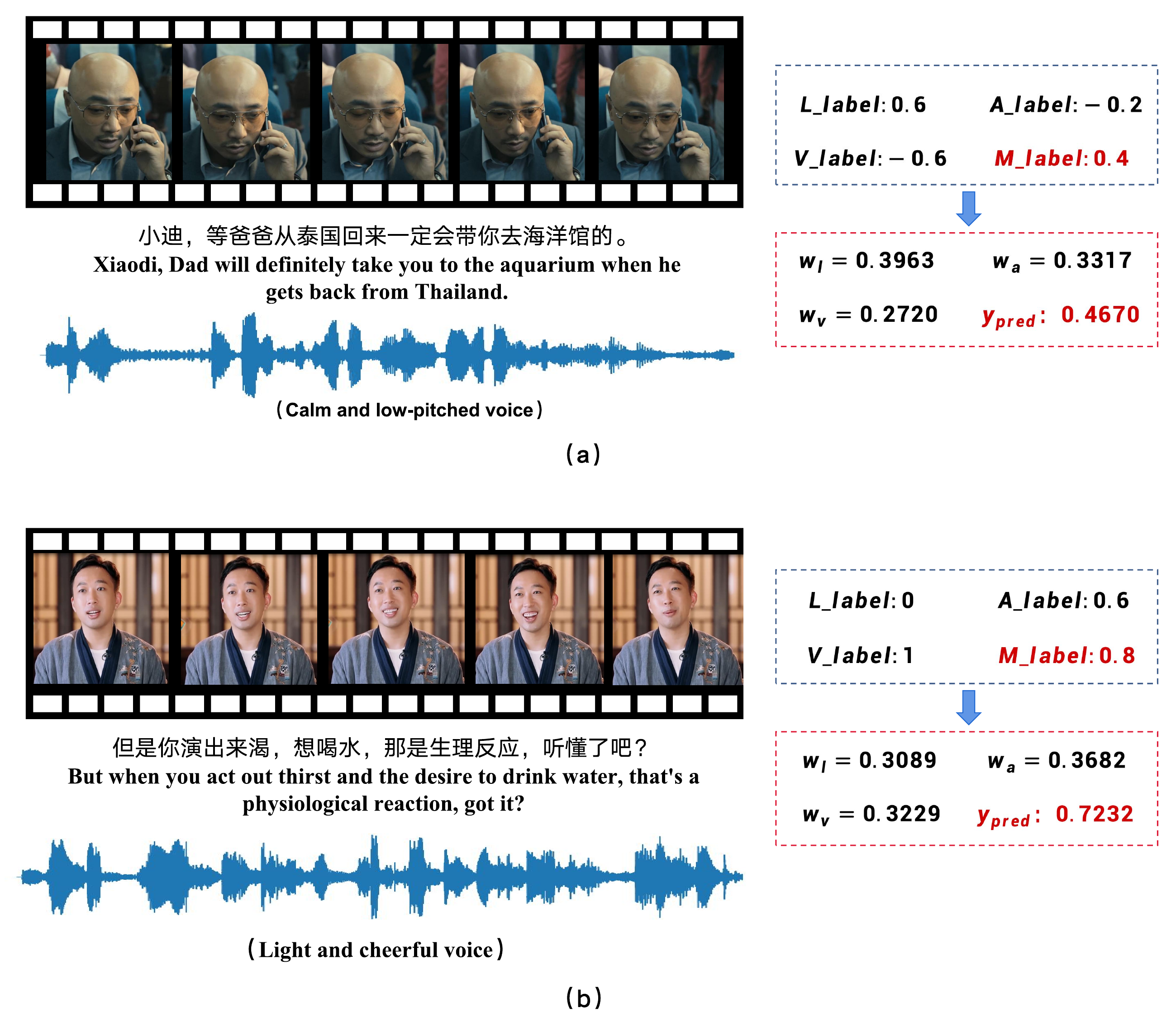}
\caption{Display of cases and modality weights on SIMS.} 
\label{case_study}
\end{figure}

\subsection{Case Study}

We illustrate MODS’s dynamic modality selection on SIMS cases with strong cross-modal conflict.
In Figure~\ref{case_study}(a), language (“aquarium”) conveys positivity, while audio-visual cues suggest negativity. MODS emphasizes language, aligning with the weakly positive label.
In Figure~\ref{case_study}(b), neutral language contrasts with positive acoustic-visual cues. MODS assigns higher weights to non-language inputs, correctly predicting positive sentiment.

\section{Conclusion}
In the MSA task, different modalities within the same sample may express inconsistent emotions. Dynamically selecting the dominant modality is crucial for accurately determining the overall sentiment. To address this, we propose the MODS algorithm, which provides a more versatile MSA solution by compressing redundant non-language modality sequences, dynamically selecting the primary modality, and progressively enhancing the primary modality. Extensive experiments demonstrate the rationality of our work.

\noindent \textbf{Future Work}. We will evaluate MODS in domains where cross-modal emotional mismatches are common, such as humor and sarcasm detection.
Meanwhile, the algorithm's robustness in real-world scenarios will also be considered, such as handling modality data gaps~\cite{yang2024TSIF,yang2024asynchronous} and spurious correlations~\cite{yang2024robust,yang2024MCIS}.

\section{Acknowledgments}
We sincerely thank Yuxuan Lei and Yue Jiang for their outstanding contributions.
\bibliography{aaai2026}

\begin{thebibliography}{53}
\providecommand{\natexlab}[1]{#1}

\bibitem[{Delbrouck et~al.(2020)Delbrouck, Tits, Brousmiche, and Dupont}]{delbrouck2020transformer}
Delbrouck, J.-B.; Tits, N.; Brousmiche, M.; and Dupont, S. 2020.
\newblock A transformer-based joint-encoding for emotion recognition and sentiment analysis.
\newblock \emph{arXiv preprint arXiv:2006.15955}.

\bibitem[{Drus and Khalid(2019)}]{drus2019sentiment}
Drus, Z.; and Khalid, H. 2019.
\newblock Sentiment analysis in social media and its application: Systematic literature review.
\newblock \emph{Procedia Computer Science}, 161: 707--714.

\bibitem[{Gallagher, Furey, and Curran(2019)}]{gallagher2019application}
Gallagher, C.; Furey, E.; and Curran, K. 2019.
\newblock The application of sentiment analysis and text analytics to customer experience reviews to understand what customers are really saying.
\newblock \emph{International Journal of Data Warehousing and Mining (IJDWM)}, 15(4): 21--47.

\bibitem[{Han et~al.(2021)Han, Chen, Gelbukh, Zadeh, Morency, and Poria}]{han2021bi}
Han, W.; Chen, H.; Gelbukh, A.; Zadeh, A.; Morency, L.-p.; and Poria, S. 2021.
\newblock Bi-bimodal modality fusion for correlation-controlled multimodal sentiment analysis.
\newblock In \emph{Proceedings of the 2021 international conference on multimodal interaction}, 6--15.

\bibitem[{Han, Chen, and Poria(2021)}]{han2021improving}
Han, W.; Chen, H.; and Poria, S. 2021.
\newblock Improving multimodal fusion with hierarchical mutual information maximization for multimodal sentiment analysis.
\newblock \emph{arXiv preprint arXiv:2109.00412}.

\bibitem[{Hazarika, Zimmermann, and Poria(2020)}]{hazarika2020misa}
Hazarika, D.; Zimmermann, R.; and Poria, S. 2020.
\newblock Misa: Modality-invariant and-specific representations for multimodal sentiment analysis.
\newblock In \emph{Proceedings of the 28th ACM International Conference on Multimedia}, 1122--1131.

\bibitem[{Hochreiter and Schmidhuber(1997)}]{hochreiter1997long}
Hochreiter, S.; and Schmidhuber, J. 1997.
\newblock Long short-term memory.
\newblock \emph{Neural Computation}, 9(8): 1735--1780.

\bibitem[{Kingma and Ba(2014)}]{kingma2014adam}
Kingma, D.~P.; and Ba, J. 2014.
\newblock Adam: A method for stochastic optimization.
\newblock \emph{arXiv preprint arXiv:1412.6980}.

\bibitem[{Kipf and Welling(2016)}]{kipf2016semi}
Kipf, T.~N.; and Welling, M. 2016.
\newblock Semi-supervised classification with graph convolutional networks.
\newblock \emph{{arXiv preprint arXiv:1609.02907}}.

\bibitem[{Lei et~al.(2023)Lei, Yang, Li, Wang, Chen, and Zhang}]{lei2023text}
Lei, Y.; Yang, D.; Li, M.; Wang, S.; Chen, J.; and Zhang, L. 2023.
\newblock Text-oriented modality reinforcement network for multimodal sentiment analysis from unaligned multimodal sequences.
\newblock In \emph{CAAI International Conference on Artificial Intelligence}, 189--200. Springer.

\bibitem[{Li, Wang, and Cui(2023)}]{li2023decoupled}
Li, Y.; Wang, Y.; and Cui, Z. 2023.
\newblock Decoupled Multimodal Distilling for Emotion Recognition.
\newblock In \emph{Proceedings of the IEEE/CVF Conference on Computer Vision and Pattern Recognition}, 6631--6640.

\bibitem[{Li et~al.(2022)Li, Zhou, Zhang, Liu, Yang, Lian, and Hu}]{li2022amoa}
Li, Z.; Zhou, Y.; Zhang, W.; Liu, Y.; Yang, C.; Lian, Z.; and Hu, S. 2022.
\newblock AMOA: Global acoustic feature enhanced modal-order-aware network for multimodal sentiment analysis.
\newblock In \emph{Proceedings of the 29th International Conference on Computational Linguistics}, 7136--7146.

\bibitem[{Liang et~al.(2021)Liang, Lin, Feng, Zhang, and Lv}]{liang2021attention}
Liang, T.; Lin, G.; Feng, L.; Zhang, Y.; and Lv, F. 2021.
\newblock Attention is not enough: Mitigating the distribution discrepancy in asynchronous multimodal sequence fusion.
\newblock In \emph{Proceedings of the IEEE/CVF International Conference on Computer Vision}, 8148--8156.

\bibitem[{Lin and Hu(2022)}]{lin2022multimodal}
Lin, R.; and Hu, H. 2022.
\newblock Multimodal contrastive learning via uni-modal coding and cross-modal prediction for multimodal sentiment analysis.
\newblock \emph{arXiv preprint arXiv:2210.14556}.

\bibitem[{Lin and Hu(2023)}]{lin2023dynamically}
Lin, R.; and Hu, H. 2023.
\newblock Dynamically shifting multimodal representations via hybrid-modal attention for multimodal sentiment analysis.
\newblock \emph{IEEE Transactions on Multimedia}, 26: 2740--2755.

\bibitem[{Liu et~al.(2022)Liu, Yuan, Mao, Liang, Yang, Qiu, Cheng, Li, Xu, and Gao}]{liu2022make}
Liu, Y.; Yuan, Z.; Mao, H.; Liang, Z.; Yang, W.; Qiu, Y.; Cheng, T.; Li, X.; Xu, H.; and Gao, K. 2022.
\newblock Make acoustic and visual cues matter: Ch-sims v2. 0 dataset and av-mixup consistent module.
\newblock In \emph{Proceedings of the 2022 international conference on multimodal interaction}, 247--258.

\bibitem[{Liu et~al.(2018)Liu, Shen, Lakshminarasimhan, Liang, Zadeh, and Morency}]{liu2018efficient}
Liu, Z.; Shen, Y.; Lakshminarasimhan, V.~B.; Liang, P.~P.; Zadeh, A.; and Morency, L.-P. 2018.
\newblock Efficient low-rank multimodal fusion with modality-specific factors.
\newblock \emph{arXiv preprint arXiv:1806.00064}.

\bibitem[{Lv et~al.(2021)Lv, Chen, Huang, Duan, and Lin}]{lv2021progressive}
Lv, F.; Chen, X.; Huang, Y.; Duan, L.; and Lin, G. 2021.
\newblock Progressive modality reinforcement for human multimodal emotion recognition from unaligned multimodal sequences.
\newblock In \emph{Proceedings of the IEEE/CVF Conference on Computer Vision and Pattern Recognition}, 2554--2562.

\bibitem[{Ma, Zhang, and Sun(2023)}]{ma2023multimodal}
Ma, F.; Zhang, Y.; and Sun, X. 2023.
\newblock Multimodal sentiment analysis with preferential fusion and distance-aware contrastive learning.
\newblock In \emph{2023 IEEE International Conference on Multimedia and Expo (ICME)}, 1367--1372. IEEE.

\bibitem[{Mai et~al.(2022)Mai, Zeng, Zheng, and Hu}]{mai2022hybrid}
Mai, S.; Zeng, Y.; Zheng, S.; and Hu, H. 2022.
\newblock Hybrid contrastive learning of tri-modal representation for multimodal sentiment analysis.
\newblock \emph{IEEE Transactions on Affective Computing}, 14(3): 2276--2289.

\bibitem[{Mao et~al.(2022)Mao, Yuan, Xu, Yu, Liu, and Gao}]{mao2022m}
Mao, H.; Yuan, Z.; Xu, H.; Yu, W.; Liu, Y.; and Gao, K. 2022.
\newblock M-SENA: An integrated platform for multimodal sentiment analysis.
\newblock \emph{arXiv preprint arXiv:2203.12441}.

\bibitem[{Ortis, Farinella, and Battiato(2019)}]{ortis2019overview}
Ortis, A.; Farinella, G.~M.; and Battiato, S. 2019.
\newblock An Overview on Image Sentiment Analysis: Methods, Datasets and Current Challenges.
\newblock \emph{ICETE (1)}, 296--306.

\bibitem[{Paszke et~al.(2019)Paszke, Gross, Massa, Lerer, Bradbury, Chanan, Killeen, Lin, Gimelshein, Antiga et~al.}]{paszke2019pytorch}
Paszke, A.; Gross, S.; Massa, F.; Lerer, A.; Bradbury, J.; Chanan, G.; Killeen, T.; Lin, Z.; Gimelshein, N.; Antiga, L.; et~al. 2019.
\newblock Pytorch: An imperative style, high-performance deep learning library.
\newblock \emph{Advances in Neural Information Processing Systems}, 32.

\bibitem[{Sabour, Frosst, and Hinton(2017)}]{sabour2017dynamic}
Sabour, S.; Frosst, N.; and Hinton, G.~E. 2017.
\newblock Dynamic routing between capsules.
\newblock \emph{Advances in neural information processing systems}, 30.

\bibitem[{Sun et~al.(2020)Sun, Sarma, Sethares, and Liang}]{sun2020learning}
Sun, Z.; Sarma, P.; Sethares, W.; and Liang, Y. 2020.
\newblock Learning relationships between text, audio, and video via deep canonical correlation for multimodal language analysis.
\newblock In \emph{Proceedings of the AAAI Conference on Artificial Intelligence}, volume~34, 8992--8999.

\bibitem[{Tsai and Wang(2021)}]{tsai2021analyzing}
Tsai, M.~H.; and Wang, Y. 2021.
\newblock Analyzing Twitter data to evaluate people’s attitudes towards public health policies and events in the era of COVID-19.
\newblock \emph{International Journal of Environmental Research and Public Health}, 18(12): 6272.

\bibitem[{Tsai et~al.(2019)Tsai, Bai, Liang, Kolter, Morency, and Salakhutdinov}]{tsai2019multimodal}
Tsai, Y.-H.~H.; Bai, S.; Liang, P.~P.; Kolter, J.~Z.; Morency, L.-P.; and Salakhutdinov, R. 2019.
\newblock Multimodal transformer for unaligned multimodal language sequences.
\newblock In \emph{Proceedings of the conference. Association for Computational Linguistics. Meeting}, volume 2019, 6558. NIH Public Access.

\bibitem[{Vaswani et~al.(2017)Vaswani, Shazeer, Parmar, Uszkoreit, Jones, Gomez, Kaiser, and Polosukhin}]{vaswani2017attention}
Vaswani, A.; Shazeer, N.; Parmar, N.; Uszkoreit, J.; Jones, L.; Gomez, A.~N.; Kaiser, {\L}.; and Polosukhin, I. 2017.
\newblock Attention is all you need.
\newblock \emph{Advances in Neural Information Processing Systems}, 30.

\bibitem[{Wang et~al.(2023{\natexlab{a}})Wang, Guo, Tian, Liu, He, and Luo}]{wang2023tetfn}
Wang, D.; Guo, X.; Tian, Y.; Liu, J.; He, L.; and Luo, X. 2023{\natexlab{a}}.
\newblock TETFN: A text enhanced transformer fusion network for multimodal sentiment analysis.
\newblock \emph{Pattern Recognition}, 136: 109259.

\bibitem[{Wang et~al.(2022)Wang, Liu, Wang, Tian, He, and Gao}]{wang2022cross}
Wang, D.; Liu, S.; Wang, Q.; Tian, Y.; He, L.; and Gao, X. 2022.
\newblock Cross-modal Enhancement Network for Multimodal Sentiment Analysis.
\newblock \emph{IEEE Transactions on Multimedia}.

\bibitem[{Wang et~al.(2023{\natexlab{b}})Wang, Li, Liang, Morency, Bell, and Lai}]{wang2023cross}
Wang, Y.; Li, Y.; Liang, P.~P.; Morency, L.-P.; Bell, P.; and Lai, C. 2023{\natexlab{b}}.
\newblock Cross-attention is not enough: Incongruity-aware dynamic hierarchical fusion for multimodal affect recognition.
\newblock \emph{arXiv preprint arXiv:2305.13583}.

\bibitem[{Wu et~al.(2023)Wu, Dai, Qin, Liu, Lin, Cao, and Sui}]{wu2023denoising}
Wu, S.; Dai, D.; Qin, Z.; Liu, T.; Lin, B.; Cao, Y.; and Sui, Z. 2023.
\newblock Denoising Bottleneck with Mutual Information Maximization for Video Multimodal Fusion.
\newblock \emph{arXiv preprint arXiv:2305.14652}.

\bibitem[{Wu et~al.(2021)Wu, Lin, Zhao, Qin, and Zhu}]{wu2021text}
Wu, Y.; Lin, Z.; Zhao, Y.; Qin, B.; and Zhu, L.-N. 2021.
\newblock A text-centered shared-private framework via cross-modal prediction for multimodal sentiment analysis.
\newblock In \emph{Findings of the Association for Computational Linguistics: ACL-IJCNLP 2021}, 4730--4738.

\bibitem[{Xiao, Yang, and Ning(2021)}]{xiao2021research}
Xiao, X.; Yang, J.; and Ning, X. 2021.
\newblock Research on multimodal emotion analysis algorithm based on deep learning.
\newblock In \emph{Journal of Physics: Conference Series}, volume 1802, 032054. IOP Publishing.

\bibitem[{Yang et~al.(2023{\natexlab{a}})Yang, Chen, Wang, Wang, Li, Liu, Zhao, Huang, Dong, Zhai, and Zhang}]{yang2023context}
Yang, D.; Chen, Z.; Wang, Y.; Wang, S.; Li, M.; Liu, S.; Zhao, X.; Huang, S.; Dong, Z.; Zhai, P.; and Zhang, L. 2023{\natexlab{a}}.
\newblock Context De-Confounded Emotion Recognition.
\newblock In \emph{Proceedings of the IEEE/CVF Conference on Computer Vision and Pattern Recognition}, 19005--19015.

\bibitem[{Yang et~al.(2022{\natexlab{a}})Yang, Huang, Kuang, Du, and Zhang}]{yang2022disentangled}
Yang, D.; Huang, S.; Kuang, H.; Du, Y.; and Zhang, L. 2022{\natexlab{a}}.
\newblock Disentangled Representation Learning for Multimodal Emotion Recognition.
\newblock In \emph{Proceedings of the 30th ACM International Conference on Multimedia}, 1642--1651.

\bibitem[{Yang et~al.(2022{\natexlab{b}})Yang, Huang, Liu, and Zhang}]{yang2022contextual}
Yang, D.; Huang, S.; Liu, Y.; and Zhang, L. 2022{\natexlab{b}}.
\newblock Contextual and Cross-Modal Interaction for Multi-Modal Speech Emotion Recognition.
\newblock \emph{IEEE Signal Processing Letters}, 29: 2093--2097.

\bibitem[{Yang et~al.(2022{\natexlab{c}})Yang, Huang, Wang, Liu, Zhai, Su, Li, and Zhang}]{yang2022emotion}
Yang, D.; Huang, S.; Wang, S.; Liu, Y.; Zhai, P.; Su, L.; Li, M.; and Zhang, L. 2022{\natexlab{c}}.
\newblock Emotion Recognition for Multiple Context Awareness.
\newblock In \emph{Proceedings of the European Conference on Computer Vision}, volume 13697, 144--162.

\bibitem[{Yang et~al.(2022{\natexlab{d}})Yang, Kuang, Huang, and Zhang}]{yang2022learning}
Yang, D.; Kuang, H.; Huang, S.; and Zhang, L. 2022{\natexlab{d}}.
\newblock Learning Modality-Specific and -Agnostic Representations for Asynchronous Multimodal Language Sequences.
\newblock In \emph{Proceedings of the 30th ACM International Conference on Multimedia}, 1708–1717.

\bibitem[{Yang et~al.(2024{\natexlab{a}})Yang, Kuang, Yang, Li, and Zhang}]{yang2024TSIF}
Yang, D.; Kuang, H.; Yang, K.; Li, M.; and Zhang, L. 2024{\natexlab{a}}.
\newblock Towards Asynchronous Multimodal Signal Interaction and Fusion via Tailored Transformers.
\newblock \emph{IEEE Signal Processing Letters}.

\bibitem[{Yang et~al.(2024{\natexlab{b}})Yang, Li, Qu, Yang, Zhai, Wang, and Zhang}]{yang2024asynchronous}
Yang, D.; Li, M.; Qu, L.; Yang, K.; Zhai, P.; Wang, S.; and Zhang, L. 2024{\natexlab{b}}.
\newblock Asynchronous Multimodal Video Sequence Fusion via Learning Modality-Exclusive and-Agnostic Representations.
\newblock \emph{IEEE Transactions on Circuits and Systems for Video Technology}.

\bibitem[{Yang et~al.(2024{\natexlab{c}})Yang, Li, Xiao, Liu, Yang, Chen, Wang, Zhai, Li, and Zhang}]{yang2024MCIS}
Yang, D.; Li, M.; Xiao, D.; Liu, Y.; Yang, K.; Chen, Z.; Wang, Y.; Zhai, P.; Li, K.; and Zhang, L. 2024{\natexlab{c}}.
\newblock Towards Multimodal Sentiment Analysis Debiasing via Bias Purification.
\newblock In \emph{Proceedings of the European Conference on Computer Vision (ECCV)}.

\bibitem[{Yang et~al.(2023{\natexlab{b}})Yang, Liu, Huang, Li, Zhao, Wang, Yang, Wang, Zhai, and Zhang}]{yang2023target}
Yang, D.; Liu, Y.; Huang, C.; Li, M.; Zhao, X.; Wang, Y.; Yang, K.; Wang, Y.; Zhai, P.; and Zhang, L. 2023{\natexlab{b}}.
\newblock Target and source modality co-reinforcement for emotion understanding from asynchronous multimodal sequences.
\newblock \emph{Knowledge-Based Systems}, 265: 110370.

\bibitem[{Yang et~al.(2024{\natexlab{d}})Yang, Yang, Li, Wang, Wang, and Zhang}]{yang2024robust}
Yang, D.; Yang, K.; Li, M.; Wang, S.; Wang, S.; and Zhang, L. 2024{\natexlab{d}}.
\newblock Robust emotion recognition in context debiasing.
\newblock In \emph{Proceedings of the IEEE/CVF Conference on Computer Vision and Pattern Recognition (CVPR)}, 12447--12457.

\bibitem[{Yu et~al.(2020)Yu, Xu, Meng, Zhu, Ma, Wu, Zou, and Yang}]{yu2020ch}
Yu, W.; Xu, H.; Meng, F.; Zhu, Y.; Ma, Y.; Wu, J.; Zou, J.; and Yang, K. 2020.
\newblock Ch-sims: A chinese multimodal sentiment analysis dataset with fine-grained annotation of modality.
\newblock In \emph{Proceedings of the 58th Annual Meeting of the Association for Computational Linguistics}, 3718--3727.

\bibitem[{Yu et~al.(2021)Yu, Xu, Yuan, and Wu}]{yu2021learning}
Yu, W.; Xu, H.; Yuan, Z.; and Wu, J. 2021.
\newblock Learning modality-specific representations with self-supervised multi-task learning for multimodal sentiment analysis.
\newblock In \emph{Proceedings of the AAAI conference on artificial intelligence}, 10790--10797.

\bibitem[{Zadeh et~al.(2017)Zadeh, Chen, Poria, Cambria, and Morency}]{zadeh2017tensor}
Zadeh, A.; Chen, M.; Poria, S.; Cambria, E.; and Morency, L.-P. 2017.
\newblock Tensor fusion network for multimodal sentiment analysis.
\newblock \emph{arXiv preprint arXiv:1707.07250}.

\bibitem[{Zadeh et~al.(2018{\natexlab{a}})Zadeh, Liang, Mazumder, Poria, Cambria, and Morency}]{zadeh2018memory}
Zadeh, A.; Liang, P.~P.; Mazumder, N.; Poria, S.; Cambria, E.; and Morency, L.-P. 2018{\natexlab{a}}.
\newblock Memory fusion network for multi-view sequential learning.
\newblock In \emph{Proceedings of the AAAI Conference on Artificial Intelligence}, volume~32.

\bibitem[{Zadeh et~al.(2016)Zadeh, Zellers, Pincus, and Morency}]{zadeh2016mosi}
Zadeh, A.; Zellers, R.; Pincus, E.; and Morency, L.-P. 2016.
\newblock Mosi: multimodal corpus of sentiment intensity and subjectivity analysis in online opinion videos.
\newblock \emph{arXiv preprint arXiv:1606.06259}.

\bibitem[{Zadeh et~al.(2018{\natexlab{b}})Zadeh, Liang, Poria, Cambria, and Morency}]{zadeh2018multimodal}
Zadeh, A.~B.; Liang, P.~P.; Poria, S.; Cambria, E.; and Morency, L.-P. 2018{\natexlab{b}}.
\newblock Multimodal language analysis in the wild: Cmu-mosei dataset and interpretable dynamic fusion graph.
\newblock In \emph{Proceedings of the 56th Annual Meeting of the Association for Computational Linguistics}, 2236--2246.

\bibitem[{Zeng et~al.(2023)Zeng, Mai, Yan, and Hu}]{zeng2023multimodal}
Zeng, Y.; Mai, S.; Yan, W.; and Hu, H. 2023.
\newblock Multimodal reaction: Information modulation for cross-modal representation learning.
\newblock \emph{IEEE Trans. Multimedia}.

\bibitem[{Zeng et~al.(2024)Zeng, Yan, Mai, and Hu}]{zeng2024disentanglement}
Zeng, Y.; Yan, W.; Mai, S.; and Hu, H. 2024.
\newblock Disentanglement Translation Network for multimodal sentiment analysis.
\newblock \emph{Inf. Fusion}, 102: 102031.

\bibitem[{Zhang et~al.(2023)Zhang, Wang, Yin, Liu, Liu, and Yu}]{zhang2023learning}
Zhang, H.; Wang, Y.; Yin, G.; Liu, K.; Liu, Y.; and Yu, T. 2023.
\newblock Learning language-guided adaptive hyper-modality representation for multimodal sentiment analysis.
\newblock \emph{arXiv preprint arXiv:2310.05804}.

\end{thebibliography}

\end{document}